\documentclass{article}
\usepackage{graphicx} 
\usepackage{amsmath}
\usepackage{amssymb}
\usepackage{bm}        
\usepackage{mathtools} 
\DeclareMathOperator{\rank}{rank}
\title{Observability Analysis of Joint Steering and Extrinsic Calibration\\[0.4em]\large Technical Report}
\author{Subodh Mishra}
\begin{document}
\maketitle

\begin{abstract}
This technical report studies the local weak observability of a planar bicycle-model vehicle when vehicle pose, planar LiDAR extrinsic calibration, and steering-angle bias are estimated jointly. A Lie-derivative-based nonlinear observability analysis is used to examine stationary, straight-line, constant-curvature, and combined straight-plus-arc motion. The resulting observability matrices and nullspaces describe how pose, LiDAR translation and yaw offsets, and steering bias become coupled under different motion primitives. Stationary motion and individual motion primitives retain unobservable directions, whereas the combination of straight and curved motion removes the identified degeneracies and yields full local weak observability of the seven-state system. The analysis provides a theoretical basis for selecting calibration trajectories that sufficiently excite both steering and sensor-extrinsic parameters.
\end{abstract}

\section{Introduction}
Joint calibration of vehicle parameters and sensor extrinsics is useful when a mobile robot must estimate both its motion state and slowly varying calibration quantities from the same measurements. In this setting, observability depends not only on the sensing model but also on the motion executed by the vehicle. Motion primitives that appear informative for navigation may still leave combinations of pose, sensor extrinsics, or steering bias indistinguishable.

This report considers a planar bicycle-model vehicle equipped with a LiDAR whose planar translation and yaw extrinsics are included in the state together with a steering-angle bias. The analysis is based on the nonlinear observability matrix constructed from the measurement Jacobian and successive Lie derivatives, following the classical differential-geometric framework for nonlinear local observability introduced by Hermann and Krener~\cite{hermann1977nonlinear}. The objective is to characterize the unobservable directions associated with stationary, straight-line, and constant-curvature motion, and then to examine how combining motion primitives changes the rank of the joint observability matrix.

The derivations below are retained explicitly so that the dependence of each result on the state, steering input, and motion primitive remains visible. The report first states the full observability matrix, then evaluates its rank and nullspace under representative motion cases, and finally summarizes the implications for calibration-trajectory design.

\section{Problem Setup and Full Observability Matrix}


\[
x = \begin{bmatrix}
x_j \\ y_j \\ \theta_j \\
x_{\mathrm{off}} \\ y_{\mathrm{off}} \\ \theta_{\mathrm{off}} \\ b
\end{bmatrix}.
\]

Shorthands:
\[
\Delta = \theta_j - \theta_i,\qquad
\phi   = \theta_{\mathrm{off}} - \theta_i,
\qquad
\alpha = \delta_m - b.
\]

Extrinsic combinations:
\[
A = \sin\Delta \, x_{\mathrm{off}} + \cos\Delta \, y_{\mathrm{off}},\qquad
B = -\cos\Delta \, x_{\mathrm{off}} + \sin\Delta \, y_{\mathrm{off}}.
\]

Trigonometric abbreviations:
\[
c_\Delta = \cos\Delta,\quad s_\Delta = \sin\Delta,\qquad
c_\phi   = \cos\phi,\quad s_\phi   = \sin\phi,
\]
\[
c_\theta = \cos(\theta_j+\phi),\qquad
s_\theta = \sin(\theta_j+\phi).
\]

The full nonlinear observability matrix is

\[
\mathcal{O}(x) =
\begin{bmatrix}
H_0(x) \\[4pt]
H_1(x) \\[4pt]
H_2(x)
\end{bmatrix}
\in \mathbb{R}^{9\times 7}.
\]


\[
H_0(x) =
\begin{bmatrix}
c_\phi &
-s_\phi &
s_\Delta x_{\mathrm{off}} + c_\Delta y_{\mathrm{off}} &
1 - c_\Delta &
s_\Delta &
-\,s_\phi (x_j - x_i) - c_\phi (y_j - y_i) &
0 \\[4pt]

s_\phi &
c_\phi &
-\,c_\Delta x_{\mathrm{off}} + s_\Delta y_{\mathrm{off}} &
-\,s_\Delta &
1 - c_\Delta &
c_\phi (x_j-x_i) - s_\phi (y_j-y_i) &
0 \\[4pt]

0 & 0 & 1 & 0 & 0 & 0 & 0
\end{bmatrix}.
\]


\[
H_1(x) =
\begin{bmatrix}
0 & 0 &
-\Big(v s_\theta + \tfrac{v}{L}B\tan\alpha\Big) &
\tfrac{v}{L}\tan\alpha\,s_\Delta &
\tfrac{v}{L}\tan\alpha\,c_\Delta &
-\,v s_\theta &
-\tfrac{v}{L}A\,\sec^2\alpha
\\[6pt]

0 & 0 &
v c_\theta + \tfrac{v}{L}A\tan\alpha &
-\,\tfrac{v}{L}\tan\alpha\,c_\Delta &
\tfrac{v}{L}\tan\alpha\,s_\Delta &
v c_\theta &
-\tfrac{v}{L}B\,\sec^2\alpha
\\[6pt]

0 & 0 & 0 & 0 & 0 & 0 & -\tfrac{v}{L}\sec^2\alpha
\end{bmatrix}.
\]


\[
H_2(x) =
\begin{bmatrix}
\nabla (L_f^2 h_x) \\[3pt]
\nabla (L_f^2 h_y) \\[3pt]
\mathbf{0}_{1\times 7}
\end{bmatrix}
\]

\[
\nabla (L_f^{2} h_x)
=
\begin{alignedat}{2}
\big[&0,\; 0,\;
-\big(\tfrac{v^{2}}{L}\tan\alpha\,c_\theta
      + \tfrac{v^{2}}{L^{2}}\tan^{2}\alpha\,A\big),
&
\;\tfrac{v^{2}}{L^{2}}\tan^{2}\alpha\,c_\Delta,
\\[4pt]
&-\tfrac{v^{2}}{L^{2}}\tan^{2}\alpha\,s_\Delta,\;
-\tfrac{v^{2}}{L}\tan\alpha\,c_\theta,
&
-\big(\tfrac{v^{2}}{L}\sec^{2}\alpha\,s_\theta
      + 2\tfrac{v^{2}}{L^{2}}\tan\alpha\sec^{2}\alpha\,B\big)
\big].
\end{alignedat}
\]

\[
\nabla (L_f^{2} h_y)
=
\begin{alignedat}{2}
\big[&
0,\; 0,\;
\tfrac{v^{2}}{L}\tan\alpha\,s_\theta
+\tfrac{v^{2}}{L^{2}}\tan^{2}\alpha\,A,
&
\;\;-\tfrac{v^{2}}{L^{2}}\tan^{2}\alpha\,s_\Delta,
\\[4pt]
&
\tfrac{v^{2}}{L^{2}}\tan^{2}\alpha\,c_\Delta,\;
\tfrac{v^{2}}{L}\tan\alpha\,s_\theta,
&
\;\;\tfrac{v^{2}}{L}\sec^{2}\alpha\,c_\theta
+2\tfrac{v^{2}}{L^{2}}\tan\alpha\sec^{2}\alpha\,A
\big].
\end{alignedat}
\]

\section{Nonlinear Observability Analysis}

We now examine the local weak observability of the joint state
\[
x=\begin{bmatrix}
x_j & y_j & \theta_j & x_{\mathrm{off}} & y_{\mathrm{off}} &
\theta_{\mathrm{off}} & b
\end{bmatrix}^\top \in\mathbb{R}^7,
\]
where $(x_j,y_j,\theta_j)$ denotes the vehicle pose, $(x_{\mathrm{off}},y_{\mathrm{off}},\theta_{\mathrm{off}})$ denotes the planar LiDAR extrinsic calibration, and $b$ denotes the steering-angle bias. The measurement is the LiDAR odometry pose expressed with respect to the reference frame $(x_i,y_i,\theta_i)$. The corresponding measurement model is
\begin{align}
h_x &= (1-\cos\Delta)\,x_{\mathrm{off}} + \sin\Delta\,y_{\mathrm{off}}
 \;+\;\cos\phi\,(x_j-x_i) - \sin\phi\,(y_j-y_i),\\
h_y &= -\sin\Delta\,x_{\mathrm{off}} + (1-\cos\Delta)\,y_{\mathrm{off}}
 \;+\;\sin\phi\,(x_j-x_i) + \cos\phi\,(y_j-y_i),\\
h_\theta &= \theta_j - \theta_i,
\end{align}
where $\Delta = \theta_j - \theta_i$ and $\phi = \theta_{\mathrm{off}} - \theta_i$. The vehicle dynamics are represented by the continuous-time bicycle model
\begin{align}
\dot x_j &= v\cos\theta_j,\qquad
\dot y_j = v\sin\theta_j,\qquad
\dot\theta_j = \frac{v}{L}\tan(\delta_m - b),
\end{align}
with $\dot x_{\mathrm{off}}=\dot y_{\mathrm{off}}
=\dot\theta_{\mathrm{off}}=\dot b=0$.

Let $f(x)$ denote the vector field defined above. For each output $h_k\in\{h_x,h_y,h_\theta\}$, the observability construction uses the output gradient together with the gradients of its first and second Lie derivatives:
\[
\mathcal{O} =
\begin{bmatrix}
\nabla h_x \\[2pt]
\nabla h_y \\
\nabla h_\theta \\[2pt]
\nabla L_f h_x \\
\nabla L_f h_y \\
\nabla L_f h_\theta \\
\nabla L_f^2 h_x \\
\nabla L_f^2 h_y \\
\nabla L_f^2 h_\theta
\end{bmatrix},
\]
This matrix collects the differential information contributed by the direct measurement, its first Lie derivatives $L_f h_k$, and its second Lie derivatives $L_f^2 h_k$. Evaluating $\mathcal{O}$ under different motion primitives reveals which combinations of the seven states can or cannot be distinguished locally.

\subsection*{1) Straight-Line Motion: $\delta_m = b$}
For the straight-line case $\delta_m = b$, the two independent nullspace
directions of $\mathcal{O}_{\mathrm{straight}}$ are
\begin{align}
v_1 &=
\begin{bmatrix}
v_{1,x} \\[2pt]
v_{1,y} \\[2pt]
0 \\[2pt]
1 \\[2pt]
0 \\[2pt]
0 \\[2pt]
0
\end{bmatrix},
&
v_2 &=
\begin{bmatrix}
v_{2,x} \\[2pt]
v_{2,y} \\[2pt]
0 \\[2pt]
0 \\[2pt]
1 \\[2pt]
0 \\[2pt]
0
\end{bmatrix},
\end{align}
with
\begin{align}
v_{1,x}
&=
\frac{
\cos(\theta_i - \theta_j)
- \cos\!\big(2\theta_i - 2\theta_{\mathrm{off}}\big)
+ \cos\!\big(\theta_i + \theta_j - 2\theta_{\mathrm{off}}\big)
- 1
}{
2\cos(\theta_i - \theta_{\mathrm{off}})
},
\\[4pt]
v_{1,y}
&=
-\sin(\theta_i - \theta_{\mathrm{off}})
+ \sin(\theta_j - \theta_{\mathrm{off}}),
\\[8pt]
v_{2,x}
&=
\frac{
\sin(\theta_i - \theta_j)
+ \sin\!\big(2\theta_i - 2\theta_{\mathrm{off}}\big)
- \sin\!\big(\theta_i + \theta_j - 2\theta_{\mathrm{off}}\big)
}{
2\cos(\theta_i - \theta_{\mathrm{off}})
},
\\[4pt]
v_{2,y}
&=
-\cos(\theta_i - \theta_{\mathrm{off}})
+ \cos(\theta_j - \theta_{\mathrm{off}}).
\end{align}
The first two components of these basis vectors depend on trigonometric functions of $\theta_i$, $\theta_j$, and $\theta_{\mathrm{off}}$. Both nullspace directions couple the vehicle position $(x_j,y_j)$ with the translational extrinsics $(x_{\mathrm{off}},y_{\mathrm{off}})$. Consequently, during straight-line motion, changes in vehicle position cannot be uniquely separated from changes in the LiDAR translation offsets, resulting in two unobservable gauge-like directions.

\subsubsection*{2) Constant-Curvature Motion (Arc): $v=v_c,\,\delta_m=\delta_c\neq b$}
For the constant-curvature (arc) case with $v=v_c$ and $\delta_m=\delta_c\neq b$,
the nullspace of $\mathcal{O}_{\mathrm{arc}}$ is one-dimensional. A basis
vector can be written as
\begin{align}
v_{\mathrm{arc}} &=
\begin{bmatrix}
v_{\mathrm{arc},x} \\[2pt]
v_{\mathrm{arc},y} \\[2pt]
0 \\[2pt]
-\dfrac{L\cos\theta_{\mathrm{off}}}{\tan(b-\delta_c)} \\[8pt]
-\dfrac{L\sin\theta_{\mathrm{off}}}{\tan(b-\delta_c)} \\[8pt]
1 \\[2pt]
0
\end{bmatrix},
\end{align}
where
\begin{align}
v_{\mathrm{arc},x}
&=
\frac{
-\,L\sin\theta_i\,\tan b\,\tan\delta_c\,\tan(\theta_i-\theta_{\mathrm{off}})
- L\sin\theta_i\,\tan(\theta_i-\theta_{\mathrm{off}})
}{\tan b - \tan\delta_c}
\nonumber\\
&\quad
+\frac{
-\,L\cos\theta_j\,\tan b\,\tan\delta_c
- L\cos\theta_j
}{\tan b - \tan\delta_c}
\nonumber\\
&\quad
+\frac{
L\cos\theta_{\mathrm{off}}\,\tan b\,\tan\delta_c
+ L\cos\theta_{\mathrm{off}}
}{
(\tan b - \tan\delta_c)\,\cos(\theta_i-\theta_{\mathrm{off}})
}
\nonumber\\
&\quad
+\frac{
-\,y_i\tan b + y_i\tan\delta_c + y_j\tan b - y_j\tan\delta_c
}{\tan b - \tan\delta_c},
\\[6pt]
v_{\mathrm{arc},y}
&=
\frac{
L\sin\theta_i\,\tan b\,\tan\delta_c
+ L\sin\theta_i
- L\sin\theta_j\,\tan b\,\tan\delta_c
- L\sin\theta_j
}{\tan b - \tan\delta_c}
\nonumber\\
&\quad
+\frac{
x_i\tan b - x_i\tan\delta_c - x_j\tan b + x_j\tan\delta_c
}{\tan b - \tan\delta_c}.
\end{align}
This nullspace direction couples the vehicle position $(x_j,y_j)$, the translational extrinsics $(x_{\mathrm{off}},y_{\mathrm{off}})$, and the extrinsic yaw $\theta_{\mathrm{off}}$. The steering bias $b$ does not participate in this nullspace direction and is therefore observable under the stated constant-curvature motion. A single arc thus removes the steering-bias ambiguity while retaining one geometric gauge direction involving pose and LiDAR extrinsics.

\subsubsection*{3) Stationary Case: $v=0$}
When the vehicle is stationary, the motion-dependent information vanishes. In this case,
\[
\rank(\mathcal{O}_{v=0}) = 3,
\]
leaving a $4$-dimensional nullspace. In addition to the two translational gauge modes that also appear during straight-line motion, two further unobservable modes remain:
\begin{itemize}
\item An extrinsic yaw mode involving $\theta_{\mathrm{off}}$, indicating that
no rotation is observed when the robot does not move.
\item A steering-bias mode (basis vector with entry $1$ in the component of $b$),
since $\dot\theta_j=0$ regardless of $b$ when $v=0$.
\end{itemize}
Thus, without motion, neither the full set of extrinsic parameters nor the steering bias can be uniquely recovered. A convenient basis for the four-dimensional nullspace of $\mathcal{O}_{v=0}$ is
\begin{align}
v_1 &=
\begin{bmatrix}
v_{1,x} \\[2pt]
v_{1,y} \\[2pt]
0 \\[2pt]
1 \\[2pt]
0 \\[2pt]
0 \\[2pt]
0
\end{bmatrix},
&
v_2 &=
\begin{bmatrix}
v_{2,x} \\[2pt]
v_{2,y} \\[2pt]
0 \\[2pt]
0 \\[2pt]
1 \\[2pt]
0 \\[2pt]
0
\end{bmatrix},
\\[4pt]
v_3 &=
\begin{bmatrix}
-\,y_i + y_j \\[2pt]
x_i - x_j \\[2pt]
0 \\[2pt]
0 \\[2pt]
0 \\[2pt]
1 \\[2pt]
0
\end{bmatrix},
&
v_4 &=
\begin{bmatrix}
0 \\[2pt]
0 \\[2pt]
0 \\[2pt]
0 \\[2pt]
0 \\[2pt]
0 \\[2pt]
1
\end{bmatrix},
\end{align}
with
\begin{align}
v_{1,x}
&=
\frac{
\cos(\theta_i - \theta_j)
- \cos\!\big(2\theta_i - 2\theta_{\mathrm{off}}\big)
+ \cos\!\big(\theta_i + \theta_j - 2\theta_{\mathrm{off}}\big)
- 1
}{
2\cos(\theta_i - \theta_{\mathrm{off}})
},
\\[4pt]
v_{1,y}
&=
-\sin(\theta_i - \theta_{\mathrm{off}})
+ \sin(\theta_j - \theta_{\mathrm{off}}),
\\[8pt]
v_{2,x}
&=
\frac{
\sin(\theta_i - \theta_j)
+ \sin\!\big(2\theta_i - 2\theta_{\mathrm{off}}\big)
- \sin\!\big(\theta_i + \theta_j - 2\theta_{\mathrm{off}}\big)
}{
2\cos(\theta_i - \theta_{\mathrm{off}})
},
\\[4pt]
v_{2,y}
&=
-\cos(\theta_i - \theta_{\mathrm{off}})
+ \cos(\theta_j - \theta_{\mathrm{off}}).
\end{align}
The first two vectors $(v_1,v_2)$ correspond to translational gauge directions that couple $(x_j,y_j)$ and $(x_{\mathrm{off}},y_{\mathrm{off}})$. The third vector $v_3$ captures the unobservable extrinsic-yaw direction, while the fourth vector $v_4$ shows that the steering bias $b$ is completely unobservable when $v=0$.

\subsubsection*{4) Combined Trajectory: Straight + Arc}
To assess the benefit of combining motion primitives, we stack the observability matrices associated with straight-line and constant-curvature motion:
\[
\mathcal{O}_{\mathrm{combined}}
=
\begin{bmatrix}
\mathcal{O}_{\mathrm{straight}} \\[4pt]
\mathcal{O}_{\mathrm{arc}}
\end{bmatrix}.
\]
The straight-motion nullspace vectors contain no component in $\theta_{\mathrm{off}}$, whereas the arc-motion nullvector has a non-zero $\theta_{\mathrm{off}}$ component. Their nullspaces therefore intersect only in the zero vector. Explicit computation gives
\[
\rank(\mathcal{O}_{\mathrm{combined}})=7,
\qquad
\mathrm{null}(\mathcal{O}_{\mathrm{combined}})=\{0\},
\]
showing that {\em all seven states become locally weakly observable} when both straight and curved segments are included in the trajectory.

\section{Conclusion}
This report examined the local weak observability of the seven-state planar bicycle-model system defined above, with LiDAR extrinsic calibration and steering-angle bias estimated jointly. The analysis shows that stationary, straight-line, and constant-curvature motion each retain characteristic unobservable directions. These degeneracies have a direct geometric interpretation: limited excitation creates ambiguity between vehicle pose and sensor extrinsics, while the steering bias becomes identifiable only when the motion provides sufficient steering-dependent information.

The straight-line and constant-curvature cases leave different nullspace directions. When both motion primitives are represented in the calibration trajectory, the combined observability matrix has full rank according to the derivation above. The result motivates calibration trajectories that deliberately mix straight and curved segments rather than relying on a single repetitive motion pattern.

The analysis is structural and local: it identifies which state directions are observable under the stated model and motion assumptions. In practical estimation systems, numerical conditioning, measurement noise, model mismatch, and the amount of excitation will additionally influence calibration accuracy and convergence.

\end{document}